\documentclass[letterpaper, 10 pt, conference]{ieeeconf}  

\newcommand\copyrighttext{%
	\footnotesize This work has been submitted to the IEEE for possible publication. Copyright may be transferred without notice, after which this version may no longer be accessible.
}
\newcommand\copyrightnotice{%
	\begin{tikzpicture}[remember picture,overlay]
	\node[anchor=south,yshift=10pt, xshift=10pt] at (current page.south) {\fbox{\parbox{\dimexpr\textwidth-\fboxsep-\fboxrule\relax}{\copyrighttext}}};
	\end{tikzpicture}%
}

\usepackage{amsmath,amssymb,bm}
\newcommand{\vectsymb}[1]{\bm{#1}}
\newcommand{\vect}[1]{\bm{#1}}        
\newcommand{\DP}[2]{\left\langle #1, #2 \right\rangle}

\usepackage{array} 
\usepackage{graphicx}
\usepackage{tikz}
\usepackage{makecell}
\usepackage{hyperref}
\usepackage{url}
\usepackage{multirow}   
\usepackage{multicol}   
\usepackage{booktabs}   
\usepackage{graphicx}   
\usepackage[table]{xcolor}
\setcellgapes{5pt} 
\usepackage{booktabs}
\usepackage{multirow}
\usepackage{pifont}
\usepackage{amsmath}
\usepackage[utf8]{inputenc}
\usepackage{longtable}
\usepackage{caption}
\usepackage{threeparttable}
\usepackage[table]{xcolor} 

\definecolor{LivingThingsBase}{RGB}{192, 201, 238}
\definecolor{VehiclesAndInfrastructureBase}{RGB}{222, 232, 206}
\definecolor{IndoorBase}{RGB}{203, 220, 235}
\definecolor{KitchenItemsBase}{RGB}{245, 210, 210}
\definecolor{SportsBase}{RGB}{189, 227, 195}
\definecolor{ElectronicsBase}{RGB}{245, 239, 230}
\definecolor{PersonalAccessoriesBase}{RGB}{232, 223, 202}

\definecolor{lightblue}{RGB}{48,92,222}
\definecolor{lightorange}{RGB}{250,156,28}

\colorlet{LivingThings}{LivingThingsBase!40}
\colorlet{VehiclesAndInfrastructure}{VehiclesAndInfrastructureBase!40}
\colorlet{Indoor}{IndoorBase!40}
\colorlet{KitchenItems}{KitchenItemsBase!40}
\colorlet{Sports}{SportsBase!40}
\colorlet{Electronics}{ElectronicsBase!40}
\colorlet{PersonalAccessories}{PersonalAccessoriesBase!40}

\usepackage{glossaries}
\newacronym{hog}{HOG}{Histograms of Oriented Gradients}
\newacronym{owod}{OWOD}{Open World Object Detection}
\newacronym{ood}{OOD}{Out-of-Distribution}
\newacronym{detr}{DETR}{DEtection TRansformer}
\newacronym{ddetr}{D-DETR}{Deformable DETR}
\newacronym{vlm}{VLM}{Vision-Language Model}
\newacronym{ovd}{OVD}{Open-Vocabulary Object Detection}
\newacronym{nlp}{NLP}{Natural Language Processing}

\title{\LARGE \bf
Beyond Flat Unknown Labels in Open-World Object Detection
}

\author{
\parbox{4in}{\centering
Yuchen Zhang, Yao Lu, Johannes Betz\\
AVS Lab, Technical University of Munich, Munich, Germany
}
}
\begin{document}

\maketitle
\copyrightnotice
\thispagestyle{empty}
\pagestyle{empty}

\begin{abstract}
Most object detectors operate under a closed-world assumption, recognizing only the classes annotated in the training dataset and failing when encountering novel objects. Open-World Object Detection (OWOD) relaxes this assumption by enabling unseen objects to be detected as \textit{Unknown}. However, collapsing all novel objects into a single undifferentiated label eliminates semantic granularity and limits informed decision-making. In this paper, we introduce BOUND, an open-world detector that advances OWOD by inferring coarse-grained categories of unknown objects rather than merely flagging their existence. 
This enriched representation offers semantic cues that may benefit real-world systems.
For example, in autonomous driving, distinguishing between an \textit{Unknown Animal} (requiring yielding) and an \textit{Unknown Debris} (requiring rerouting) leads to fundamentally different planning behaviors. Technically, BOUND integrates a sparsemax-based head for modeling objectness, a hierarchy-guided relabeling component that provides auxiliary supervision, and a classification module that learns hierarchical relationships. Experiments on OWOD benchmarks demonstrate that BOUND achieves higher unknown recall than existing baselines without sacrificing known-class mAP, while additionally enabling structured hierarchical categorization of unknown instances. Furthermore, evaluations on the long-tail LVIS dataset demonstrate robust generalization. Code is available at: \url{https://anonymous.4open.science/r/BOUND}

\end{abstract}

\section{Introduction}

Object detection, which is a critical task in computer vision that involves both localizing and classifying objects, has evolved significantly over the years. Early approaches relied on hand-crafted features \cite{viola, HOG}, but were later replaced by deep learning methods, which can be broadly categorized into one-stage detectors (e.g., YOLO~\cite{yolo}) and two-stage detectors (e.g., Faster R-CNN~\cite{fasterrcnn}). More recently, vision-transformer-based detectors, pioneered by \gls{detr} \cite{detr}, have gained significant popularity. By leveraging the attention mechanisms in transformers, these models are able to capture long-range dependencies and global context.

Despite these advancements, a challenge persists: most detectors can only recognize the classes defined in their training datasets \cite{OWOD, OWDETR}. This stems from the \textit{closed-world assumption}, which presumes that all objects encountered during deployment are already seen and learned during training. Such an assumption is not only unrealistic but can also pose safety risks in practical applications. For instance, an autonomous vehicle trained on pedestrians and cars would fail to detect \gls{ood} objects (e.g., e-scooters or construction barriers), causing unsafe behavior.

To address this safety gap, recent research on \gls{owod} \cite{OWOD, 2bpcd, OWDETR, PROB, luo2024opendetd, xi2024ktcn} has focused on detecting \gls{ood} objects and label them as \textit{Unknown}. However, simply localizing unknown objects is not sufficient. When encountering something unfamiliar, humans rarely assign a generic \textit{Unknown} label but instead categorize it into broader, semantically meaningful groups. For instance, at a fruit market, an unfamiliar object resembling an orange would more likely be regarded as an \textit{Unknown Fruit} rather than just an \textit{Unknown Object}. Applying this idea to object detection can be highly beneficial. If an autonomous vehicle detects a deer ahead as an \textit{Unknown Animal}, the system can infer that the object might exhibit potential movement and therefore wait. In contrast, if a construction barrier is detected and categorized as an \textit{Unknown Obstacle}, the vehicle should instead generate a bypass trajectory, since the object is expected to remain stationary.

Building on this idea, we propose incorporating a hierarchical taxonomy of objects to derive more information of unknown objects. The proposed framework, BOUND, is designed to localize both known and unknown objects. Known objects are classified into their fine-grained categories, represented by the leaf nodes of the taxonomy, while unknown objects are assigned to coarser categories at higher levels of the hierarchy.

Our main contributions are as follows:
\begin{itemize}
    \vspace{-3pt}
    \item  We extend the standard \gls{owod} setting by introducing the task of categorizing unknown objects into meaningful coarse categories, rather than treating them as a single undifferentiated class.
    \item  We present BOUND, which integrates (i) a sparsemax-based objectness head for predicting whether a bounding box encloses a valid object, independent of category, (ii) a hierarchy-aware classification module enforcing taxonomic consistency, and (iii) hierarchy-guided relabeling for auxiliary supervision of objectness.
    \item Experiments show that BOUND improves unknown recall without sacrificing known-class mAP, supports hierarchical categorization, and demonstrates strong generalization.
\end{itemize}

\section{Related Work}

\noindent \textbf{DETR-based Detectors and OWOD.} \hspace{0.3cm}
As the first fully end-to-end object detector, \gls{detr} \cite{detr} reformulated detection as a set prediction problem using a CNN backbone and a transformer, eliminating hand-crafted components such as anchor generation and non-maximum suppression.
\gls{ddetr} \cite{DDETR} improved the original framework by introducing multi-scale adaptive sparse sampling to address convergence and small object localization.
Subsequent variants introduced further architectural improvements \cite{ConditionalDETR, DABDETR, DINOdetr}. In parallel, \gls{owod} considers scenarios where novel objects appear at test time as \textit{Unknowns} and are later incorporated as known classes once the labels are given.
Methods such as OW-DETR \cite{OWDETR} and PROB \cite{PROB} use objectness scores or regression heads to localize unknowns. More recent research utilizes foundation models like SAM \cite{SAM} and CLIP \cite{radford2021clip} to provide pseudo-labels or prototypes for unknown discovery \cite{he2024SGROD, xi2024ktcn, luo2024opendetd}. However, these primarily focus on localization, whereas our work enables hierarchical categorization of unknown objects.

\noindent \textbf{Open-Vocabulary Object Detection.} \hspace{0.3cm}
\gls{ovd} extends object detection beyond fixed label sets by conditioning on text prompts using \glspl{vlm} such as CLIP \cite{radford2021clip}. Existing methods mainly follow two paradigms: (i) knowledge distillation, which aligns region features with text embeddings to bridge the localization gap \cite{gu2021RKD, tarpn, wang2023oadp, hdovd}, and (ii) self-training, which leverages zero-shot \gls{vlm} predictions to generate pseudo labels and expand the vocabulary \cite{PBOVD, VLPLM}.
Despite these advances, OVD differs fundamentally from our setting: it relies on an uncontrolled vocabulary learned during \gls{vlm} pretraining and assumes the availability of text prompts at inference, whereas our task requires detecting unknown objects without any semantic specification.

\noindent \textbf{Hierarchical Classification.}\hspace{0.3cm}
Hierarchical classification organizes categories into a taxonomy to exploit semantic relationships, evolving from traditional machine learning methods \cite{EW1,wordnet2,EW3} to deep models that explicitly enforce hierarchical constraints \cite{HEX,HDCNN}. However, most approaches are limited to known classes and cannot generalize to novel categories. To address this, Lee et al. \cite{Lee} introduced strategies that allow predictions to stop at higher-level nodes to indicate novelty, later refined by evidence-based formulations to better separate known and unknown classes \cite{Evidence}. Despite these advances, hierarchical classification was studied almost exclusively in the image classification, whereas we extend it to object detection.
\section{Preliminaries}

\noindent \textbf{D-DETR.}\hspace{0.3cm}
We adopt \gls{ddetr} \cite{DDETR} as the base detection architecture. It consists of four main components: 
(1) a CNN backbone that extracts multi-scale feature maps from the input image; 
(2) a transformer encoder–decoder module, where the encoder processes these features and the decoder uses a fixed set of learned object queries with multi-scale deformable attention to produce object-level embeddings; 
(3) prediction heads that map each query embedding to a class probability and a bounding box; and 
(4) a Hungarian matching mechanism used during training to assign each ground-truth object to a unique query, with unmatched queries treated as background.

\noindent \textbf{Sparsemax.}\hspace{0.3cm}
Sparsemax was originally proposed by Martins et al.\cite{sparsemax} in the context of \gls{nlp} tasks as an alternative to the softmax function, particularly for multi-label text classification and sequence modeling. Unlike softmax, which always produces dense distributions, sparsemax projects scores onto the probability simplex and can assign exactly zero probability to irrelevant classes, yielding more compact and interpretable outputs. Subsequent works \cite{sparseattention0,sparseattention1,sparseattention3} adopted sparsemax in attention mechanisms, where its sparsity encourages more selective and interpretable alignments while preserving performance comparable to softmax. However, to the best of the our knowledge, sparsemax has not been employed as the activation function in the final layer of computer vision models, where softmax or sigmoid activations remain standard.

\begin{figure}[htbp]
\centering
\includegraphics[width=0.6\linewidth]{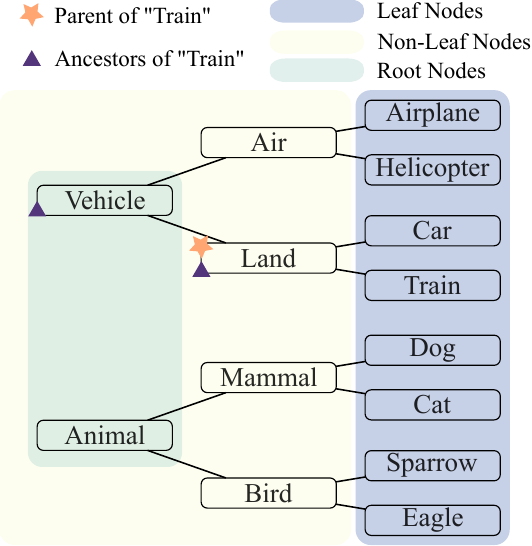}
\caption{Example taxonomy structure.}
\label{fig:hie_example}
\end{figure}

\noindent \textbf{Hierarchical Taxonomy.}\hspace{0.3cm}
We represent the label space as a directed forest $\mathcal{T} = (N, E)$, where $N$ is the set of nodes, covering the complete hierarchy from coarse-grained (non-leaf) to fine-grained (leaf) classes, and an edge $(a, b) \in E$ indicates that class $a$ is a direct hypernym of class $b$. 
For any node $c \in N$, its \emph{parent} is denoted by $p(c)$, where $(p(c), c) \in E$. The set of \emph{ancestor nodes} is defined as $\mathrm{Anc}(c) = \{ x \in N \mid (x, c) \in E^+ \}$, where $E^+$ denotes the transitive closure of $E$. The set of \emph{leaf nodes} is $L = \{ c \in N \mid \nexists x \in N : (c, x) \in E \}$, the set of \emph{non-leaf nodes} is $P = N \setminus L$, and the set of \emph{root nodes} is $R = \{ c \in N \mid \nexists x \in N : (x, c) \in E \}$. An illustrative example is provided in Figure \ref{fig:hie_example}.

\section{Method}

\begin{figure*}[h]
\begin{center}
\includegraphics[width=.9\linewidth]{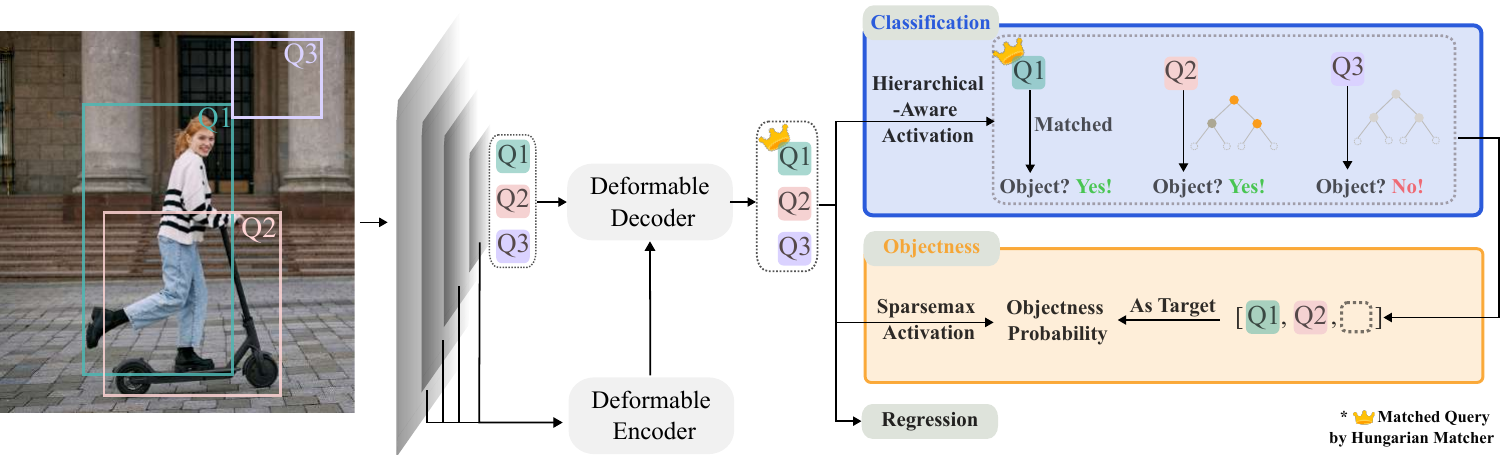} 
\end{center}
   \caption{
\textbf{Overall pipeline of the proposed method.} Building on \gls{ddetr}, the classification head first applies a hierarchy-aware activation that couples parent and child classes. Based on these activations, we apply a hierarchy-guided relabeling strategy. In standard \gls{ddetr}, only queries that are matched to ground-truth objects by the Hungarian matcher are labeled as positives, with all others treated as background. In contrast, our approach relabels queries with strong non-leaf activations as potential objects (e.g., Q2), while queries with weak activations remain background (e.g., Q3).  The target in the objectness head will be updated accordingly, providing auxiliary supervision that complements the primary objectness modeling and helps refine objectness learning. The objectness head uses sparsemax to model competition and sparsity among queries, guided by the updated labels from the classification head. Regression head remains identical as the original \gls{ddetr}. For clarity, the Hungarian matcher is omitted, and matched queries are indicated by a crown symbol.}
\label{fig:pipeline}
\end{figure*}

In this section, we present the details of our proposed method (Figure \ref{fig:pipeline}). It consists of three main components: an objectness head with sparsemax, a hierarchy-aware activation, and a hierarchy-guided relabeling strategy.

\subsection{Objectness with Sparsemax}

\noindent \textbf{Motivation.}\hspace{0.3cm}
In addition to the classification and localization heads in \gls{ddetr}, an objectness head is introduced. This head takes the embeddings from the last decoder layer and produces a scalar score for each query, with sparsemax applied as the activation. The choice of sparsemax is motivated by two considerations:

(1) \textit{Competition Rather than Suppression.} A straightforward approach would be to use a sigmoid activation with hard binary supervision for each query individually. However, this design forces unknown objects to share the negative target with background queries, and thus being suppressed. To address this, we reformulate the problem from an individual query perspective to a collective one: all queries within an image compete for objectness probability allocation. Sparsemax naturally facilitates this competition by allocating a probability budget across queries, encouraging known objects to receive higher scores, while plausible but unannotated queries are not explicitly forced to zero.

(2) \textit{Sparsity.} Sparsemax produces sparse probability distributions, i.e., many outputs are exactly zero. This aligns well with the characteristics of query-based object detectors, where most queries capture background regions, while only a few correspond to real objects. The sparsity induced by sparsemax therefore provides a principled way to reflect this underlying nature.

\noindent \textbf{Sparsemax-based Objectness.}\hspace{0.3cm}
With the motivations established, we now describe sparsemax in the objectness head. Specifically, given the logits $\vectsymb{z} \in \mathbb{R}^{Q}$ of all $Q$ queries, sparsemax projects them onto the probability simplex, yielding a sparse probability distribution $\vectsymb{p}$:

\begin{equation}
\mathrm{sparsemax}(\vectsymb{z})
= \underset{\vectsymb{p} \in \Delta^{Q-1}}{\arg\min} \;
\|\vectsymb{p} - \vectsymb{z}\|^2,
\end{equation}
where $\Delta^{Q-1}$ is the (Q - 1)-dimensional simplex:
\[
\Delta^{Q-1} = \left\{ \vectsymb{p} \in \mathbb{R}^Q
\;\middle|\; \DP{\vect{1}}{\vectsymb{p}} = 1,\;
\vectsymb{p} \ge \vect{0} \right\}.
\]

This projection produces a probability vector $\vectsymb{p}$ in which many entries are exactly zero, while the active entries remain positive and sum to one. Unlike softmax, which assigns nonzero probability to every query, sparsemax highlights only a subset of queries as relevant. As a result, the scores also become easier to interpret: non-informative background queries  are discarded, while only queries with evidence of objects retain positive values.

For training, all positive queries share the probability budget equally, and the sparsemax loss is used:

\begin{equation}\label{eq:sparsemax_loss}
L_{\mathrm{obj}} = -\vectsymb{q}^\top\vectsymb{z} + \frac{1}{2} \sum_{j \in S(\vectsymb{z})} (z_j^2 - \tau^2(\vectsymb{z})) + \frac{1}{2}\|\vectsymb{q}\|^2,
\end{equation}

 where $\vectsymb{z}$ represents the vector of logits, $\vectsymb{q}$ is the target distribution vector, $S(\vectsymb{z})$ denotes the support set, which is the subset of indices where the sparsemax output is non-zero, and $\tau(\vectsymb{z})$ is a threshold value that determines the boundary of this support set.

\subsection{Hierarchy-Aware Activation}

\noindent \textbf{Motivation.}\hspace{0.3cm}
Conventional classification heads treat categories as independent and typically return a single label without considering the hierarchy among classes. Simply adding non-leaf nodes into the vector while still treating outputs independently is insufficient. On the one hand, it neglects the natural coupling between parent and child classes, which can lead to inconsistent predictions (e.g., predicting a child but not its parent). On the other hand, enforcing strict coupling, where a child can only be active if its parent is active, removes inconsistencies but introduces error propagation: mistakes at higher levels cascade downward and prevent recovery at the leaves. For example, if a \textit{Sparrow} is misclassified into the branch \textit{Mammal} instead of \textit{Bird}, all bird subclasses become inaccessible and the correct label can never be reached. These limitations motivate explicitly encoding the hierarchy into the model during training, with greater emphasis placed on coarse-level nodes.

\noindent \textbf{Hierarchy-Aware Activation.}\hspace{0.3cm}
To achieve this, we extend the classification head with a hierarchy-aware activation function. Let $ y \in (0,1)^{k}$ denote per-class sigmoid activations. For each child class $c$ with parent $p(c)$, we define:

\begin{equation}
\tilde{y} = y_{c} \cdot (y_{p(c)})^{\alpha_{c}},
\label{equ:strength}
\end{equation}

where $\alpha_{c}$ is a learnable strength parameter. Root nodes remain unchanged ($\tilde{y}_{r} = y_{r}$). This multiplicative formulation reinforces hierarchical consistency by coupling children with their parents, while simultaneously propagating parent errors to descendants, thereby increasing the penalty associated with inaccurate coarse-level predictions and highlighting their importance.

The decision to make $\alpha_c$ learnable arises from the observation that the strength of coupling between a child and its parent is not uniform across the taxonomy. For instance, within the parent \textit{Bird}, a \textit{Sparrow} inherits most of its defining features and therefore requires strong coupling, whereas a \textit{Penguin} diverges from the typical bird shape and thus exhibits weaker coupling in a vision-based system. Assigning each child its own learnable parameter $\alpha_c$ allows the model to capture such variability and adaptively infer the appropriate coupling strength in a data-driven manner.

\subsection{Hierarchy-Guided Relabeling}

\noindent \textbf{Motivation.} \hspace{0.3cm}
Relabeling can provide auxiliary supervisory signals by leveraging the model’s own predictions rather than relying solely on annotated ground truth. The idea is to let high-confidence signals from the classification head act as provisional guidance for the objectness head. Although such pseudo-labels are less precise than human-provided annotations, they enrich supervision by reinforcing queries that appear object-like, thereby refining objectness learning.

\noindent \textbf{Relabeling.} \hspace{0.3cm}
The classification head is supervised using a class hierarchy. For queries matched to ground-truth instances via Hungarian matching, the supervision target is a multi-hot vector where both the ground-truth leaf class and all its ancestors are assigned a positive label. For unmatched queries, supervision is limited to the leaf nodes, which are all assigned a negative target. This prevents the model from classifying them into specific object categories while allowing ambiguity at higher levels of the hierarchy. In essence, it tells the model, ``We are certain this does not correspond to any specific known class, but it may belong to a broader object category''.

While the sparsemax-based objectness head plays the primary role in highlighting plausible object queries, we complement it with a lightweight relabeling strategy. Specifically, as induced by our supervision signal in the classification head, the non-leaf predictions of unmatched queries are not explicitly suppressed. As a result, unmatched queries may still exhibit meaningful confidence at non-leaf levels, which we interpret as potential evidence of unknown objects. To make use of this signal, we define an adaptive confidence threshold based on the minimum non-leaf score among all matched queries in an image. Any unmatched query whose non-leaf prediction exceeds this threshold is relabeled as a candidate unknown, and the target in the objectness loss is adjusted accordingly.

\section{Experiments}

\begin{table*}[h]
\centering
\caption{\textbf{Quantitative results on OWOD and OW-DETR Splits.} We report U-R, mAP, and HAcc, which respectively capture the localization of unknown objects, the detection of known objects, and the hierarchical categorization of unknowns. In Task 4, all classes are treated as known, so U-R and HAcc are omitted. Across both splits, BOUND achieves consistently higher U-R while maintaining a strong balance between known and unknown detection. Some methods (e.g., RandBox and ALLOW-DETR) report higher mAP, but this comes at the cost of substantially reduced U-R. Importantly, BOUND is the only model capable of categorizing unknowns into the hierarchy. N/A indicates that the corresponding functionality is not supported by the method.}
\setlength{\tabcolsep}{3pt} 
\resizebox{0.7\textwidth}{!}{%
\begin{tabular}{lcccccccccc}
\toprule
\multirow{2}{*}{Method}
& \multicolumn{3}{c}{Task 1}
& \multicolumn{3}{c}{Task 2}
& \multicolumn{3}{c}{Task 3}
& \multicolumn{1}{c}{Task 4} \\
\cmidrule(lr){2-4} \cmidrule(lr){5-7} \cmidrule(lr){8-10} \cmidrule(lr){11-11}
& U-R $\uparrow$
& mAP $\uparrow$
& HAcc $\uparrow$
& U-R $\uparrow$
& mAP $\uparrow$
& HAcc $\uparrow$
& U-R $\uparrow$
& mAP $\uparrow$
& HAcc $\uparrow$
& mAP $\uparrow$ \\
\midrule
\multicolumn{11}{l}{\textbf{OWOD Split}} \\
\midrule
ORE-EBUI \cite{OWOD} & 4.9 & 56.0 & \multirow{8}{*}{\begin{tabular}{c}\vrule height 8ex width 0.6pt\\[0ex]N/A\\[0ex]\vrule height 8ex width 0.6pt\end{tabular}} & 2.9 & 39.4 & \multirow{8}{*}{\begin{tabular}{c}\vrule height 8ex width 0.6pt\\[0ex]N/A\\[0ex]\vrule height 8ex width 0.6pt\end{tabular}} & 3.9 & 29.7 & \multirow{8}{*}{\begin{tabular}{c}\vrule height 8ex width 0.6pt\\[0ex]N/A\\[0ex]\vrule height 8ex width 0.6pt\end{tabular}} & 25.3 \\
UC-OWOD \cite{ucowod}  & 2.4 & 57.0 & & 3.7 & 31.8 & & 8.7 & 24.6 & & 23.2 \\
OCPL \cite{yu2023ocpl}  & 8.3 & 56.6 & & 7.6 & 39.1 & & 11.9 & 30.7 & & 26.7 \\
2B-OCD \cite{2bpcd}  & 10.5 & 56.4 & & 9.0 & 38.1 & & 11.6 & 29.2 & & 25.8 \\
RandBox \cite{randbox} & 10.6 & \textbf{61.8} & & 6.3 & 45.3 & & 7.8 & \textbf{39.4} & & \textbf{35.4} \\
OW-DETR \cite{OWDETR} & 7.5 & 59.2 & & 6.2 & 42.9 & & 5.7 & 30.8 & & 27.8 \\
ALLOW-DETR \cite{allowdetr} & 13.6 & 59.3 & & 10.0 & \textbf{45.6} & & 14.3 & 38.0 & & 30.6 \\
PROB \cite{PROB} & 19.4 & 59.5 & & 17.4 & 44.0 & & 19.6 & 36.0 & & 31.5 \\
\rowcolor{blue!5}
\textbf{Ours: BOUND} & \textbf{20.9} & 57.9 & \textbf{29.9} & \textbf{20.6} & 44.1 & \textbf{15.3} & \textbf{22.0} & 36.8 & \textbf{28.6} & 32.7 \\
\midrule
\multicolumn{11}{l}{\textbf{OW-DETR Split}} \\
\midrule
ORE-EBUI \cite{OWOD} & 1.5 & 61.4 & \multirow{3}{*}{\begin{tabular}{c}\vrule height 1.5ex width 0.6pt\\[0ex]N/A\\[-1ex]\vrule height 1.5ex width 0.6pt\end{tabular}} & 3.9 & 40.6 & \multirow{3}{*}{\begin{tabular}{c}\vrule height 1.5ex width 0.6pt\\[0ex]N/A\\[-1ex]\vrule height 1.5ex width 0.6pt\end{tabular}} & 3.6 & 33.7 & \multirow{3}{*}{\begin{tabular}{c}\vrule height 1.5ex width 0.6pt\\[0ex]N/A\\[-1ex]\vrule height 1.5ex width 0.6pt\end{tabular}} & 31.8 \\
OW-DETR \cite{OWDETR} & 5.7 & 71.5 & & 6.2 & 43.8 & & 6.9 & 38.5 & & 33.1 \\
PROB \cite{PROB} & 17.6 & \textbf{73.4} & & 22.3 & 50.4 & & 24.8 & 42.0 & & 39.9 \\
\rowcolor{blue!5}
\textbf{Ours: BOUND} & \textbf{22.6} & 72.7 & \textbf{5.0} & \textbf{24.8} & \textbf{52.0} & \textbf{4.0} & \textbf{28.3} & \textbf{45.8} & \textbf{5.0} & \textbf{44.4} \\
\bottomrule
\end{tabular}%
}
\label{tab:quan_main}
\end{table*}

\noindent \textbf{Datasets.}\hspace{0.3cm}
The experiments were conducted on the OWOD benchmarks, namely the OWOD Split \cite{OWOD}, and the OW-DETR Split \cite{OWDETR}. Images are taken from the PASCAL-VOC2007/2012 \cite{everingham2010pascalvoc} and MS-COCO \cite{coco} datasets, comprising a total of 80 object classes. In each split, the classes are divided into four non-overlapping subsets, each representing a separate task.  During training on a given task, only annotations for the classes belonging to that task are provided. The tasks are presented incrementally: when training on Task 2, for instance, the model must learn the new classes introduced in that task while retaining knowledge of the classes from Task 1. During evaluation, the model is tested on the entire test dataset, where images may contain objects from both previously learned classes and classes that have not yet been introduced. The two benchmarks differ in how classes are distributed across tasks: OW-DETR Split groups semantically similar categories together (e.g., all animals belong to Task 1), OWOD Split distributes classes relatively evenly. The experimental taxonomy in BOUND is derived from WordNet \cite{Wordnet} with limited human intervention to resolve category ambiguities (e.g., \emph{Toaster} may refer to an appliance or a person proposing a toast) and can be adapted to different tasks or applications.

\noindent \textbf{Metrics.}\hspace{0.3cm}
For known objects, we report the mean Average Precision (mAP), computed over all classes the model has encountered up to the current task. To assess the model's handling of unknown objects, we report two metrics:  Unknown Recall (U-R) measures how many objects from future classes are correctly detected as unknown, where in our model ``unknown'' refers to predictions that stop at a non-leaf node in the hierarchy.  Hierarchy Accuracy (HAcc) evaluates whether detected unknown objects are assigned to the correct parent node in the semantic hierarchy. Formally,
$
\text{HAcc} = \frac{1}{N} \sum_{i=1}^{N} \mathbf{1}\!\left( \hat{p}(c_i) = p(c_i) \right)
$
, where $N$ is the number of detected unknown objects, $c_i$ is the ground-truth class of the $i$-th object, $p(c_i)$ is its ground-truth parent, and $\hat{p}(c_i)$ is the predicted parent. 

\noindent \textbf{Baselines for Comparison.}\hspace{0.3cm}
We evaluate BOUND against established methods in OWOD. Our primary baselines are DETR-based approaches, namely OW-DETR \cite{OWDETR},  PROB \cite{PROB}, and ALLOW-DETR \cite{allowdetr}. Additionally, we report results from Faster R-CNN–based methods, including ORE-EBUI \cite{OWOD}, UC-OWOD \cite{ucowod}, OCPL \cite{yu2023ocpl}, 2B-OCD \cite{2bpcd}, and RandBox \cite{randbox}, as informative references.

\noindent \textbf{Implementation.}\hspace{0.3cm}
We adopt the D-DETR \cite{DDETR} architecture with a ResNet-50 backbone \cite{resnet} pretrained using the DINO self-supervised method \cite{dino}. For incremental learning, we adopt the exemplar replay strategy from PROB \cite{PROB}. The model uses 100 object queries, each with a hidden dimension of 256. Both the transformer encoder and decoder consist of six layers. Training is conducted on four NVIDIA H100 GPUs with a batch size of 8 per GPU.

\noindent \textbf{Quantitative Results.}\hspace{0.3cm}
\label{para:assump}
Table \ref{tab:quan_main} presents the quantitative results on both the OWOD and OW-DETR Splits. Across tasks, BOUND consistently improves U-R, indicating stronger ability to localize unknown objects. On OWOD, some methods (e.g., RandBox, ALLOW-DETR) achieve higher mAP but with substantially lower U-R. For example, while RandBox obtains a 6.7\% higher mAP, its unknown recall is 49.3\% lower than ours. On the OW-DETR Split, which was introduced more recently and thus has fewer reported baselines, BOUND shows steady gains in both U-R and mAP (with the only exception of Task 1 mAP).

A key strength of BOUND is its ability to assign previously unseen objects to correct parent categories, a capability missing in existing methods. On the OWOD Split, BOUND achieves up to 29.9\% HAcc, demonstrating meaningful hierarchical reasoning beyond simply separating knowns from unknowns. However, HAcc on the OW-DETR Split is considerably lower. This is mainly due to the OW-DETR Split design, which groups similar classes within the same task. As a result, the model lacks exposure to certain parent categories. For instance, if a task only includes subclasses under \textit{Animal}, the model has no chance to learn the concept of \textit{Food}. Consequently, even when BOUND successfully detects unknown objects in OW-DETR Split, their assignments to parent nodes become arbitrary.

\begin{figure*}[h]
\begin{center}
\includegraphics[width=0.85\linewidth]{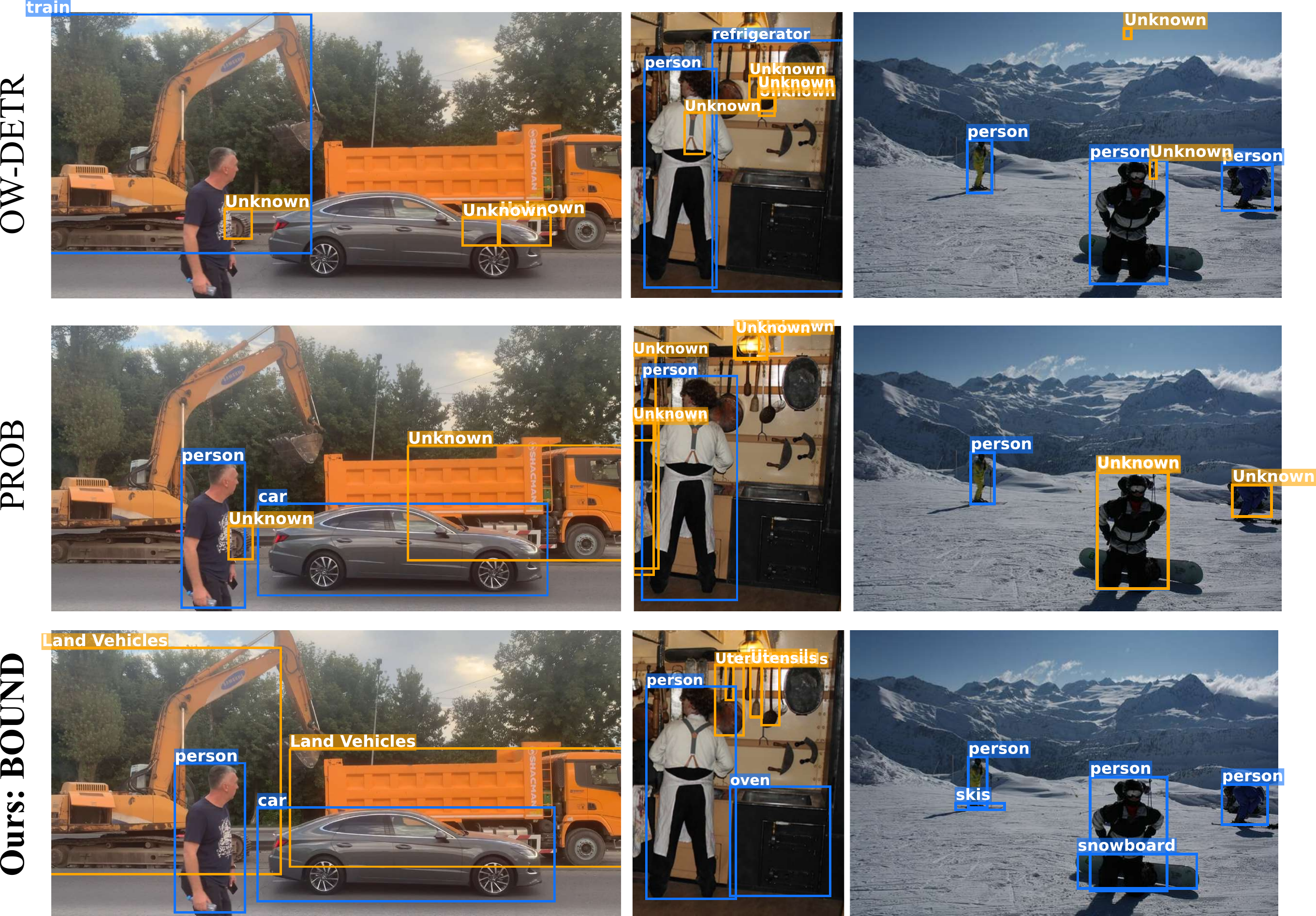} 
\end{center}
   \caption{\textbf{Qualitative results from BOUND (bottom row) compared with OW-DETR (top row) and PROB (middle row).} Predicted known objects are shown in \textcolor{lightblue}{blue}, while predicted unknown objects are shown in \textcolor{lightorange}{orange}. The first two columns illustrate BOUND’s capability to detect unknown objects: BOUND not only localizes them accurately but also assigns meaningful coarse categories (e.g., the excavator in the first image and the spatula in the second image). The third column highlights BOUND’s stable performance in detecting known objects. For fair comparison, the same number of top-k predictions is shown for each image.
   }
\label{fig:qualitative}
\end{figure*}

\noindent \textbf{Qualitative Results.}\hspace{0.3cm}
Figure \ref{fig:qualitative} presents a qualitative comparison among OW-DETR, PROB, and our proposed BOUND. OW-DETR often misses unknown objects and occasionally known ones. For instance, in the first example, both the car and the person are missed, while in the second image a refrigerator is incorrectly hallucinated. PROB achieves more reliable localization, e.g., correctly identifying the lamp in the second image, but it can also confuse known objects with unknowns, e.g., misclassifying the people in the third. In contrast, BOUND not only improves the detection of known objects, successfully capturing the oven in the second image, and the skis and the snowboard in the third image, but also categorizes unknowns into semantically meaningful groups. For example, the excavator in the first example is recognized as a \textit{Land Vehicle}, and the spatula in the second example is grouped under \textit{Utensils}. These examples highlight the capability of BOUND to move beyond generic \textit{Unknown} labels by providing structured and interpretable categorizations of novel objects, thereby enhancing both robustness and semantic richness in open-world detection.

\noindent \textbf{Scalability.}\hspace{0.3cm}
To demonstrate the scalability, we test BOUND on the LVIS dataset \cite{gupta2019lvis} ($\sim$1,200 classes), splitting classes randomly into two subsets: two-thirds as known during training and the remaining one-third as unknown during evaluation. LVIS exhibits an extreme long-tail distribution, with approximately 40\% of classes containing fewer than 50 instances across 100,170 images, which leads to inherently extremely low overall mAP. For this reason, and to enable controlled comparison with Table \ref{tab:quan_main}, we additionally report mAP-COCO, computed on the overlapping categories between COCO and LVIS within the known set (the class mapping between COCO and LVIS follows \cite{cocolvis}). As can be seen from Table \ref{tab:lvis}, PROB shows severe degradation in known object detection, whereas BOUND maintains stable performance on both known objects (mAP-COCO) and unknown objects (U-R) without hyperparameter tuning. HAcc reveals a natural trade-off: deeper hierarchies provide finer semantic granularity but reduce parent-node assignment accuracy due to increased structural complexity. Notably, the improved HAcc relative to Table~\ref{tab:quan_main} validates our earlier hypothesis regarding the low HAcc on the OW-DETR split. Unlike the OWOD benchmarks, which tend to group semantically similar categories within the same task, the random LVIS split avoids such semantic isolation and enables more robust hierarchical reasoning, resulting in higher HAcc.

\begin{table}[h!]
\centering
\caption{LVIS scalability results.}
\begin{tabular}{lcccc}
\toprule
\textbf{Method} & \textbf{mAP} & \textbf{mAP-COCO} & \textbf{U-R} & \textbf{HAcc} \\
\midrule
PROB               & 2.9  & 11.5   & 30.1   & N/A   \\
\textbf{Ours: BOUND} (depth = 3)   & 12.0   & 38.8   & 26.8   & 79.5   \\
\textbf{Ours: BOUND} (depth = 5)   & 12.0 & 38.7 & 26.3 & 38.1 \\
\bottomrule
\end{tabular}%
\label{tab:lvis}
\end{table}

\section{Ablation Study}

To assess the contribution of individual components in BOUND, we perform an ablation study on all tasks of the OWOD Split. Each variant modifies a single component of the full model, and performance is compared against the complete configuration shown in the last row of Table \ref{tab:ablation}. The following aspects are investigated:

\begin{table*}[h]
\centering
\caption{\textbf{Ablation study of key components in the BOUND model on OWOD Split.} Each row measures the performance impact of a specific modification compared to the full model (bottom): (a) \textit{w/ Softmax-Obj:} Replaces the default sparsemax activation with softmax activation in the objectness head. (b) \textit{w/o Relabel:} Disables the relabeling. (c) \textit{w/o L-Strength:} Disables the dynamic learning of hierarchical class relationships.}  
\setlength{\tabcolsep}{3pt}
\resizebox{0.8\textwidth}{!}{%
\begin{tabular}{lcccccccccc}
\toprule
\multirow{2}{*}{Configuration}
& \multicolumn{3}{c}{Task 1}
& \multicolumn{3}{c}{Task 2}
& \multicolumn{3}{c}{Task 3}
& \multicolumn{1}{c}{Task 4} \\
\cmidrule(lr){2-4} \cmidrule(lr){5-7} \cmidrule(lr){8-10} \cmidrule(lr){11-11}
& U-R $\uparrow$
& mAP $\uparrow$
& HAcc $\uparrow$
& U-R $\uparrow$
& mAP $\uparrow$
& HAcc $\uparrow$
& U-R $\uparrow$
& mAP $\uparrow$
& HAcc $\uparrow$
& mAP $\uparrow$ \\
\midrule

w/ Softmax-Obj & 18.1 & 51.7 & 24.8  & 17.3 & 35.6 & \textbf{18.9}  & 18.7 & 30.4 & 28.4  & 28.2 \\
w/o Relabel & 20.3 & 57.7 & 30.8 & 19.7 & 43.6 & 17.6 & 20.4 & 36.1 & 26.9  & 32.4\\
w/o L-Strength & 19.6 & \textbf{58.1} & \textbf{31.5}  & 18.8 & 43.9 & 16.3& 21.3 & \textbf{37.0} & \textbf{30.7} & 32.6 \\
\rowcolor{blue!5}
BOUND (Full) & \textbf{20.9} & 57.9 & 29.9  & \textbf{20.6} & \textbf{44.1} & 15.3 & \textbf{22.0} & 36.8 & 28.6  & \textbf{32.7} \\
\bottomrule
\end{tabular}
}
\label{tab:ablation}
\end{table*}

\noindent \textbf{Sparsemax Objectness.} \hspace{0.3cm}
We evaluate the effect of sparsemax in the objectness head by replacing it with softmax, following the comparison setup in \cite{sparsemax}. As shown in Table~\ref{tab:ablation}, this change leads to a consistent drop in U-R (up to 3.3\%) and mAP (up to 8.5\%).

This behavior is consistent with the different gradient characteristics of the two activations. Softmax assigns non-zero gradients to all queries, including a large number of background queries, which can dilute learning signals for true objects. In contrast, sparsemax yields sparse gradients concentrated on a subset of high-confidence queries, allowing the objectness head to focus on positive instances.

Although softmax achieves higher HAcc in Task~2, this metric is computed over the detected unknown instances. When unknown recall is lower, HAcc reflects performance on a reduced set, and should therefore be interpreted together with unknown recall.

\noindent \textbf{Relabel.} \hspace{0.3cm} 
In this ablation, we disable relabeling and use only explicitly annotated objects as positive targets for the objectness head. This results in a modest but consistent reduction in unknown recall across tasks. This indicates that although relabeling contributes to the discovery of unknown objects, the sparsemax-based objectness head serves as the primary contributor. Interestingly, a small but consistent drop in known-class mAP is also observed. This suggests that relabeling may also play a stabilizing role during training, likely by reinforcing reliable proposals and reducing the impact of missing or noisy annotations.

\noindent \textbf{Learnable Strength.}\hspace{0.3cm} 
We ablate the learnable strength $\alpha_c$ by fixing it to zero. This results in a consistent reduction in unknown recall and an increase in HAcc. Without hierarchical coupling, the model tends to follow the path of least resistance, focusing on the specific and visually distinct features of leaf-node classes while neglecting the more abstract parent categories. This bias reduces its ability to generalize, leading to lower unknown recall. At the same time, H-Acc rises, since the fewer unknowns that are detected tend to resemble known leaf classes and are therefore more easily assigned to the correct branch of the hierarchy.

\section{Future Work}

While our current relabeling method improves recall for unknown objects, many such objects still go undetected. We believe this reflects a fundamental limitation of image-based OWOD methods: because the notion of ``objectness'' is learned primarily from known classes, these methods are inherently biased toward them. Consequently, they are more likely to detect unknown objects that share visual similarity with known categories, whereas those with distinct appearances often go unnoticed. A promising future direction is to leverage Vision-Language Models (VLMs). Beyond their broad knowledge, VLMs possess a built-in understanding of semantic hierarchies, making them particularly well suited for guiding relabeling at different taxonomic levels. Incorporating such hierarchical cues could enable more structured pseudo-labels, substantially improving the identification and categorization of unknown objects.

Another future direction could be the utilization of multimodal data to strengthen open world detection. Visual cues alone may be insufficient when unfamiliar objects share limited similarity with known categories. Complementary modalities, such as audio or thermal imaging, can provide additional discriminative signals. For example, rare vehicles such as tractors or excavators can be distinguished through characteristic engine sounds, while different animals may be detected via thermal cues. Such multimodal fusion has the potential to create richer representations, enabling more reliable detection and characterization of unknown objects.

\section{Conclusion}
This work introduces BOUND, a novel and semantically rich framework for OWOD. Moving beyond the conventional \textit{Unknown} label, BOUND provides the capability to categorize novel objects into coarse classes, enabling a more informed and nuanced understanding of the open world. The framework is rooted in three strong, interconnected modules: (1) a sparsemax-based objectness head that enforces competition among queries and produces sparse, interpretable objectness scores,  (2) a hierarchy-aware activation that enforces consistency across taxonomy levels, and (3) a hierarchy-guided relabeling that leverages non-leaf activations to provide auxiliary supervision for objectness. Collectively, these three modules enable BOUND to deliver strong performance. Experiments demonstrate that BOUND improves unknown recall, enables meaningful coarse-grained categorization of unknowns, and exhibits strong generalization. By providing richer semantic information about unknown objects, this work extends OWOD beyond a simplistic known–unknown dichotomy.

\section*{ACKNOWLEDGMENT}
The initial draft of this manuscript was written by the authors. Large language models were used as an assistive tool to improve grammar and clarity. All suggested changes were reviewed by the authors, who take complete responsibility for the final version of this manuscript.

\bibliographystyle{IEEEtran}
\bibliography{reference}

\end{document}